\documentclass[10pt,twocolumn,letterpaper]{article}

\usepackage{iccv}
\usepackage{times}
\usepackage{epsfig}
\usepackage{graphicx}
\usepackage{amsmath}
\usepackage{amssymb}
\usepackage{amssymb}
\usepackage{makecell}

\usepackage[breaklinks=true,bookmarks=false]{hyperref}

\iccvfinalcopy 

\def\iccvPaperID{****} 

\ificcvfinal\fi

\title{Semantically Interpretable Activation Maps: what-where-how explanations within CNNs}

\author{Diego Marcos\\
 Wageningen University\\
The Netherlands\\
{\tt\small diego.marcos@wur.nl}
\and
Sylvain Lobry\\
 Wageningen University\\
The Netherlands\\
{\tt\small sylvain.lobry@wur.nl}
\and
Devis Tuia\\
 Wageningen University\\
The Netherlands\\
{\tt\small devis.tuia@wur.nl}
}

\newcommand{\Sy}[2]{\textcolor{red}{#2}}
\newcommand{\diego}[2]{\textcolor{magenta}{#2}}
\newcommand{\devis}[2]{\textcolor{blue}{#2}}
\begin{document}

\maketitle

\ificcvfinal\thispagestyle{empty}\fi

\begin{abstract}
A main issue preventing the use of Convolutional Neural Networks (CNN) in end user applications is the low level of transparency in the decision process. 
Previous work on CNN interpretability has mostly focused either on localizing the regions of the image that contribute to the result or on building an external model that generates plausible explanations. However, the former does not provide any semantic information and the latter does not guarantee the faithfulness of the explanation.
We propose an intermediate representation composed of multiple Semantically Interpretable Activation Maps (SIAM) indicating the presence of predefined attributes at different locations of the image. These attribute maps are then linearly combined to produce the final output. This gives the user insight into what the model has seen, where, and a final output directly linked to this information in a comprehensive and interpretable way.
We test the method on the task of landscape scenicness (aesthetic value) estimation, using an intermediate representation of 33 attributes from the SUN Attributes database.
The results confirm that SIAM makes it possible to understand what attributes in the image are contributing to the final score and where they are located. Since it is based on learning from multiple tasks and datasets, SIAM improve the explanability of the prediction without additional annotation efforts or computational overhead at inference time, while keeping good performances on both the final and intermediate tasks.  
\end{abstract}

\section{Introduction}
\label{sec:intro}
Deep learning (DL) models are nowadays entering many fields of application due to their clear advantages in terms of prediction accuracy. Among the different DL models, deep Convolutional Neural Networks (CNN) dominate the landscape of  Computer Vision tasks and keep expectations aloft by promises of superhuman autonomous driving or health diagnosis.
At the same time, a drawback of DL \Sy{, as relevant as models' performance when it comes to high stakes applications, LA PHRASE EST SUPER LONGUE SINON, ET CETTE PARTIE AMENE PAS BCP}{} is increasingly being put forward: the inscrutable nature of their decision making process. Often referred to as black boxes, CNNs don't allow to easily understand what elements of the input contributed to the output and in which way~\cite{knight2017techreview}.

\begin{figure}
    \centering
    \includegraphics[width=.9\columnwidth]{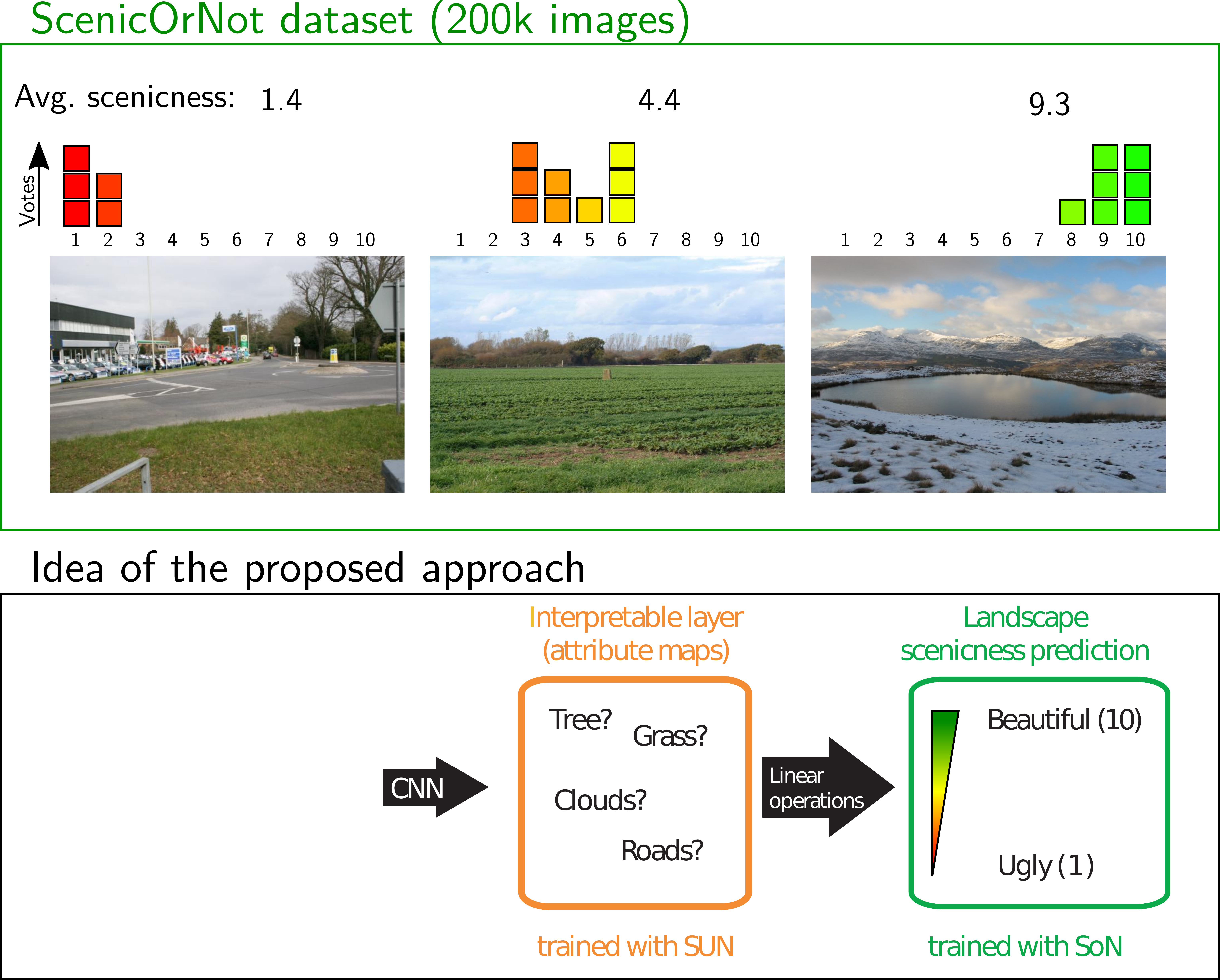}
    \caption{Examples of images from the  ScenicOrNot dataset and corresponding crowdsourced scenicness scores (top). We propose to make use of two distinct datasets such that the final task, scenicness prediction, is solved by linearly combining the results on a more interpretable intermediate task, attribute prediction using the images from the SUN Attributes database (bottom). }
    \label{fig:my_label}
\end{figure}

The end user might require an explanation that is simple enough to be easily interpretable, while the CNN needs to perform a highly complex set of operations to solve the task~\cite{gilpin2018explaining}. This creates a trade-off between how faithful the explanation is to the inner workings of the CNN and how interpretable it is~\cite{herman2017promise}.
In this paper, we argue that the explanation should actually be part of the model. This is achieved by an interpretable bottleneck ensuring that the explanation contains all the information being used to produce the result. This effectively eases the tradeoff between the interpretability of the prediction and the faithfulness of such interpretation, since the explanation will include both aspects by design. However, this risks to create  another trade-off, this time between interpretability and performance, due to the limiting capacity of the interpretable bottleneck. 

We explore the possibility to gain interpretability by learning (and predicting) an intermediate semantic representation from auxiliary datasets on related tasks. We do so by constraining the bottleneck of a CNN to predict class-specific maps, which are useful to interpret the final decision of the model on an harder to interpret final task. We rely on the idea of interpretable decomposition~\cite{zhou2018interpretable}, where we assume that the final task can be explained as a linear combination of a series of semantic contributions.

As the final task, we focus on the highly subjective visual problem of estimating landscape scenicness (i.e. aesthetic value)~\cite{seresinhe2017using,workman2017understanding}. We use a crowdsourced dataset from the  ScenicOrNot\footnote{url{http://scenicornot.datasciencelab.co.uk/}} project (SoN, Fig.~\ref{fig:my_label}, top). The model needs to capture the average perception of a large amount of annotators. This subjectiveness makes it hard for a user to evaluate the faithfulness of the prediction of such model making it important to understand which visual elements led to the final decision. To provide evidence on the model's inner decision process, we force it to use a combination of objective elements (a subset of 33 relevant SUN Attributes \cite{patterson2014sun}, Fig.~\ref{fig:my_label}, bottom) in its last intermediate representation layer, just before providing the scenicness score. By doing so, the user receives both the score and the relative contribution of each interpretable element as a set of attention maps: both can be further used to assess confidence and/or generate new knowledge about landscape preferences.

Our results suggest that it is possible to make a CNN predicting scenincness interpretable in terms of semantic landscape elements, and this without increasing the annotation or computational efforts and with a minimal decrease in terms of performance on the final task.

\section{Related work}

Interpretable deep learning for solving visual tasks is becoming a major research field. This section provides a review of the different strategies that have been explored in this direction.


\paragraph{Attribute based zero-shot learning.}

The relationship between attributes and classes is exploited for zero-shot learning, a learning setting in which some classes have no training samples, but can be related to known classes via a shared set of attributes.
Given a set of known classes $\mathcal{Y}$ with at least one training sample and a disjoint set of hidden classes $\mathcal{Z}$ (no training samples), zero-shot learning~\cite{larochelle2008zero} aims at assigning a new element from the input feature space $\mathcal{X}$ to one class in $\mathcal{Y}\cup\mathcal{Z}$. This can be done by leveraging a set of attributes $\mathcal{A}$, also referred to as Semantic Output Codes~\cite{palatucci2009zero}, common to $\mathcal{Y}$ and $\mathcal{Z}$, that can be used to uniquely describe each class~\cite{farhadi2009describing}.
Direct Attribute Prediction (DAP)~\cite{lampert2014attribute} is a family of methods for zero-shot learning in which two functions, $f: \mathcal{X}\rightarrow\mathcal{A}$ and $g: \mathcal{A}\rightarrow\mathcal{\mathcal{Y}\cup\mathcal{Z}}$ are composed to perform classification. 

Although not initially devised to improve interpretability, we propose to use an architecture inspired in DAP to make sure that the final result depends only on the learned attributes. 

\paragraph{Attributes without supervision.} Citing relevant visual attributes is an intuitive way of explaining an image-based decision. CNNs have been shown to automatically learn representations that are well correlated to visual attributes~\cite{escorcia2015relationship,bau2017network} that can be leveraged to get an intuitive idea of what elements are used for the output~\cite{harradon2018causal,olah2018building}. This behaviour can be improved further by adding a loss during training that makes the activation maps of each filter more attribute-like, such as by encouraging them to be class-specific and localized~\cite{zhang2018interpretable}. Nevertheless most individual filters in current approaches can not readily be assigned a semantic label~\cite{gonzalez2018semantic}. 

\paragraph{Interpretability by localization.} A direction to improve the interpretability of a CNN's result is to point out which parts of the input contribute the most to the output. Researches  considered occlusions of parts of the image as ways to assess the region's importance~\cite{zeiler2014visualizing,ribeiro2016should,petsiuk2018rise}. 
Alternatively, Class Activation Maps (CAM~\cite{zhou2016learning}), use an average pooling operation on the last feature tensor, right before the last fully connected layer, to assess region importance. This allows to see which locations are being used the most to generate the output. Grad-CAM~\cite{selvaraju2017grad} and LRP~\cite{montavon2018methods} use the gradient information to backtrack an output to the input elements it is most sensitive to. Such localization methods have been shown to improve the perceived trustworthiness of DL models~\cite{ribeiro2016should}

\paragraph{Interpretability by generating explanations.} Interpretbility by localization  lacks the expressivity that is expected from explanations in human communication. This has been addressed by building an external model that is trained to generate a plausible explanation to the output of the visual model~\cite{hendricks2016generating}, which can then also be combined with localization~\cite{huk2018multimodal,hendricks2018grounding}. 
Another approach to present post-hoc semantic explanations is to decompose the activation map provided by a localization method, such as CAM, using an interpretable basis of maps~\cite{zhou2018interpretable}, such that the final map is reconstructed using a combination of maps that are semantically interpretable. \diego{\devis{}{[WHY THIS LAST WORK SHOULD BE HERE? SHOULDN'T IT BE ON THE PREVIOUS PARA?]} Diego: [The previous one is about localization, the most used, and with no semantics]}{}

\paragraph{Compositionality.} Generally referred to the semantics of natural language, compositionality implies that the meaning of an expression is formed by the combination of the meanings of its parts. This principle can also be found in applications on images, such as for the task of extracting high-level information by combining low-level cues~\cite{tu2005image}, and on videos, in order to use the presence of concepts, for instance individual actions and objects, to classify video sequences as belonging to an event category (\emph{e.g.} wedding, sport event, \emph{etc.})~\cite{yu2012multimedia}. 
Exploring the high-level representation space of CNNs using sets of images containing the same concept has been proposed as a way to hint these concept's presence in the image~\cite{kim2018interpretability}.
Methods for making the flow of information in CNNs as local as possible are also researched to make the models more compositional~\cite{stone2017teaching,simpson2019gradmask}, since they encourage each individual region of the input to contribute to the output independently from its context. 

\paragraph{Joint learning of semantic hierarchies.} Tasks such as action recognition are well suited to a hierarchical representation, in which objects~\cite{kalogeiton2017joint} and object sub-actions~\cite{zhao2017single} can be learned jointly and combined to obtain the final result.  Following the same logic, images can be represented by a semantic bottleneck that describes them and that can then be used for some downstream task. The bottleneck can adopt different forms, such as pieces of text that describe the image~\cite{bucher2018semantic}, objects~\cite{daniels2018scenarionet,losch2019interpretability} or object parts and their attributes~\cite{ak2018learning}. Such representations are intelligible for humans and can thus be easily interpreted. 

\begin{figure*}[!t]
    \centering
    \includegraphics[width=1.7\columnwidth]{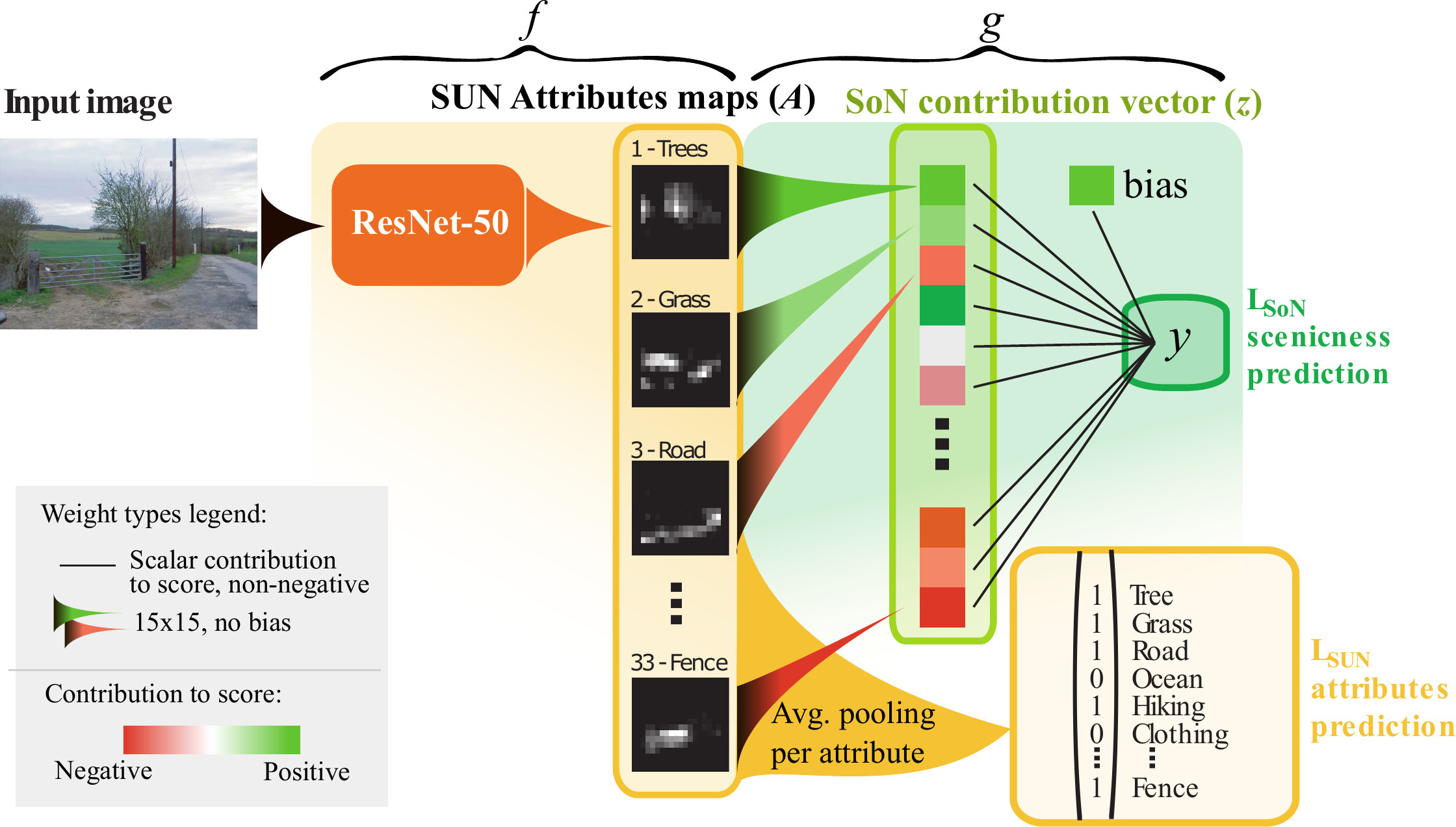}
    \caption{Flowchart of the proposed model. Each attribute map is multiplied with a learned template (see Fig.~\ref{fig:templates} for a visualization), and the resulting activations are linearly combined to obtain the scenicness score. If the input image belongs to the SUN database, only the $L_{SUN}$ loss is computed and $f$ is updated. If it belongs to the SoN database, only the $L_{SoN}$, which allows to update $f\circ g$.}
    \label{fig:flowchart}
\end{figure*}

\vspace{.3cm}

In this work, we design an interpretable layer performing localization of objects and  attributes in the image. Similarly to CAM, we use activation maps before the fully connected layers, but we force those maps to correspond to fixed concepts. We exploit the idea of compositionality by assuming that our main uncertain task (scenicness prediction) can be predicted by a linear combination of semantic interpretable concepts, which we learn in a supervised fashion. To do so, we use a dataset (SUN Attributes) disjoint from the one employed to train for the main task (SoN). In this way we exploit the semantic information contained in the auxiliary dataset and provide interpretable intermediate maps, as well as a transparent look at their importance in the final decision. 

\section{Semantically Interpretable Activation Maps (SIAM).}

SIAM consists of an end-to-end trainable CNN based on a two-level hierarchical output with a DAP~\cite{lampert2014attribute} structure. As in~\cite{zhang2018interpretable}, we want to obtain interpretable feature maps without the need for any additional annotation. But instead of relying on an unsupervised loss, we make use of an already existing dataset that contains relevant attributes (or concepts). This removes the requirement of having to inspect a substantial part of the dataset to understand the correct interpretation of each feature map, since the attributes are predefined. Our approach reassembles the interpretable basis representation method of~\cite{zhou2018interpretable}, with the main difference that our system is trained end-to-end and does not allow a residual; all the high-level information used to solve the final task must be contained in the interpretable maps. 
Our model also uses the average pooling technique of CAM~\cite{zhou2016learning} to also provide the approximate location on the image of each attribute without the need for any positional ground truth and without any additional overhead at inference time.

Figure~\ref{fig:flowchart} summarizes the proposed Semantically Interpretable Activation Maps (SIAM) architecture using as example a subset of the SUN Attributes as the semantic bottleneck $\mathcal{A}$ and landscape scenicness as final output $\mathcal{Y}$. The first block of the model, $f: \mathcal{X}\rightarrow\mathcal{A}$, outputs as many feature maps as there are attribute classes (Section~\ref{ssec:f}). This first level output is used as input to the second block, $g: \mathcal{A}\rightarrow\mathcal{Y}$, which multiplies each map with a learned spatial template (see Fig.~\ref{fig:templates} for a visualization of templates after training), and linearly combines the resulting activations to obtain the final scenicness score (Section~\ref{ssec:g}). This direct dependence between the attribute maps $\mathcal{A}$ and the final output $\mathcal{Y}$ allows to understand what elements are being detected and how they are contributing to the output. 

The two blocks are trained jointly using the corresponding datasets providing labels for attributes and scenicness, respectively (Section~\ref{ssec:tr}).

\subsection{Predict attributes with $f$}\label{ssec:f}

As $f$ we use a standard CNN architecture, ResNet-50. Given an input image $\mathbf{x}\in\mathcal{X}=\mathbb{R}^{M\times M \times 3}$, the output $f(\mathbf{x})$ is a tensor $\mathbf{A}\in\mathcal{A}=\mathbb{R}^{m\times m \times A}$ of activation maps (one map $\mathbf{A}_i\in\mathbb{R}^{m\times m}$ for each attribute $i\in \{1,\dots,A\}$). 
In practice we have used $M=500$ and $m=15$, but only a center crop of the maps of size $11\times 11$ is used, to reduce potential border effects. 
An average pooling is then applied to the maps to return a vector $\mathbf{a}\in\mathbb{R}^A$ of length $A$, the number of attribute classes. The elements $a_i \in \mathbf{a}$ are subject to a sigmoid non-linearity before being compared to the the ground truth attribute annotation $\bar{a}_i \in \bar{\mathbf{a}}$ via a multi-class binary cross entropy\Sy{,}{:}
\begin{equation}
L_{SUN} = \frac{1}{A}\sum_i [\bar{a}_i \log(a_i) + (1-\bar{a}_i) \log(1-a_i)], \label{eq:f}
\end{equation}
\Sy{if the attribute ground truth is available. THIS IS DUE TO YOUR PROCEDURE OF TRAINING}{}

\subsection{Combine attributes for the final result with $g$}\label{ssec:g}

We choose the function $g$, with the output $y=g(\mathbf{A})$ being a scalar, to be formed by a concatenation of linear operators, making the mapping between $\mathcal{A}$ and $\mathcal{Y}$ easily interpretable as a single linear mapping that provides a template for each attribute. This operation can be thought of as a weighted average pooling, where each element of the template indicates the weight of the attribute, positive or negative, towards the final output at every spatial location (see Figure~\ref{fig:templates} for the templates obtained for the SUN attribute classes).

A single fully connected template $\mathbf{T}^i$ is multiplied element-wise with each individual attribute map, yielding a single scalar per attribute $z^i=\mathbf{T}^i\cdot \mathbf{A}^i$, without using a bias term. The absence of a bias ensures that the output is non-zero exclusively if the attribute has been detected. To initialize the templates we chose not to use any randomization, since that could inject biases in the how attributes affect the final output. Instead, we learn two non-negative templates for each attribute map. Both are then combined, one with a positive weight, the other with a negative weight. This allows to initialize both templates with a constant positive value, reducing the bias and leading to a spatially smoother result. The linear combination makes it algebraically equivalent to using a single template. 
The resulting vector $\mathbf{z}\in\mathbb{R}^A$ is then linearly projected onto the output scalar $y=\mathbf{w}\cdot\mathbf{z}+b$, this time with a learnable bias, which can capture the average value of the output over the dataset. This output is then compared to the crowdsourced scenicness value $\bar{y}$ using a Square Error loss, 
\begin{equation}
L_{SoN} = (y - \bar{y})^2. \label{eq:g}
\end{equation}

\begin{figure*}[h!]
    \centering
    \includegraphics[width=1.8\columnwidth]{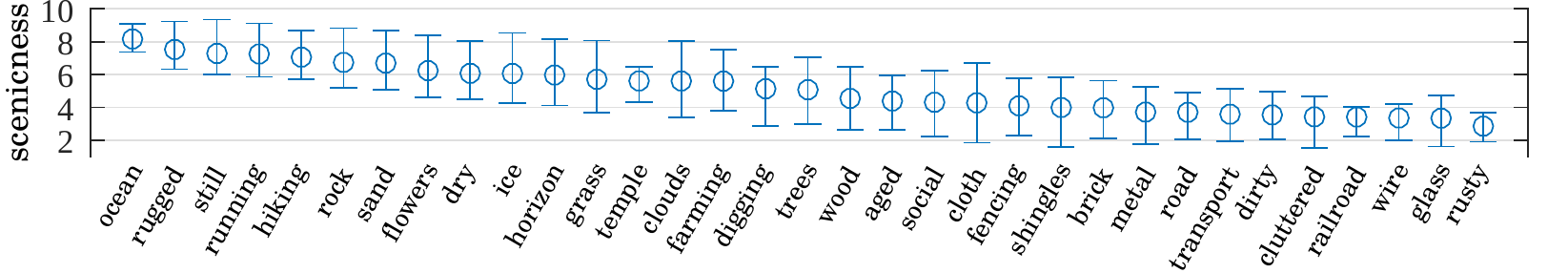}
    \caption{Co-occurrence between class presence in an image and its scenicness value on our dataset of annotated attributes over SoN images. Running and still refer to water bodies, and ice includes snow.\Sy{PUT LATEX CHAR FOR A AND Y + INPUT IMAGE -> INPUT IMAGE(x)}{}}
    \label{fig:classbeauty}
\end{figure*}

\subsection{Joint training}\label{ssec:tr}

SIAM solves two inter-dependent tasks: the prediction of attributes, $\mathbf{A}=f(\mathbf{x})$, and the prediction of scenicness based on these attributes, $y=g(\mathbf{A})$.
These two tasks involve two separate losses, needing annotations of attributes (Eq.~\eqref{eq:f}) or scenicness scores  (Eq.~\eqref{eq:g}), respectively. We use two different and non-overlapping datasets for these two tasks: the SUN Attributes database~\cite{patterson2014sun} is used to learn the attribute maps and the ScenicOrNot (SoN \footnote{\url{http://scenicornot.datasciencelab.co.uk/}}) is used as a reference for the main task of predicting the scenicness. In practice, we first fine-tune the first part of the model ($f$) on the sub-task of predicting the attributes by minimizing $L_{SUN}$. Then, the network is finetunded again using both tasks, 
\begin{equation}
L = L_{SUN}+0.1L_{SoN},
\end{equation}
in order to learn $f\circ g$ using samples from both the SUN and the SoN databases. Samples from either database are used alternately, and only one of the two losses propagates gradients at each time: when a SUN sample is considered, only $L_{SUN}$ generates a learning signal; when the sample is from SoN, only $L_{SoN}$ does. The contribution of $L_{SUN}$ is set to be an order of magnitude larger than that of $L_{SoN}$ to prevent the model from improving on scenicness prediction at the expense of its performance on the attributes.

\section{Results and discussion}

\begin{figure*}[h]
    \centering
    \includegraphics[width=1.8\columnwidth]{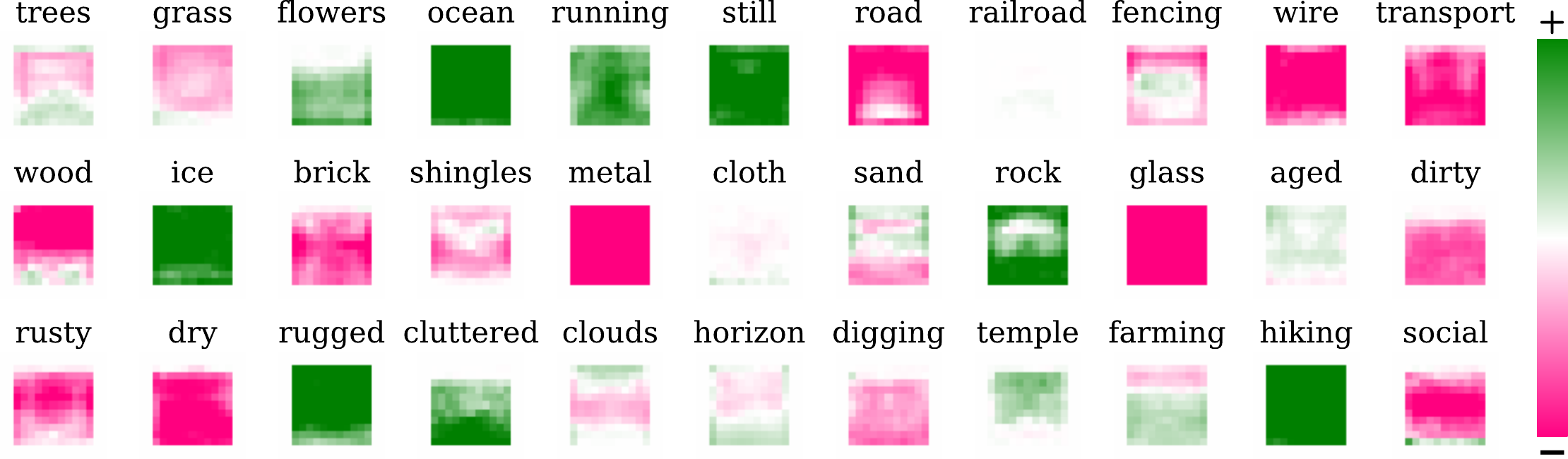}
    \caption{Learned templates on each attribute class that are used by the model for scenicness prediction. They allow to visualize how the presence of each attribute in different locations of the image influences the scenicness score. }
    \label{fig:templates}
\end{figure*}

To investigate the `performance \emph{vs.} interpretability' trade-off, we test the proposed approach on the problem of automatic landscape scenicness estimation. We use the dataset provided by the SoN project, from which we obtained the first $200'000$ listed outdoor images from the UK with at least 3 crowd-sourced scenicness scores. Ordered by ID, we took the first $150'000$ images for training, followed by $20'000$ images for validation purposes and $30'000$ for testing (Table~\ref{tab:n}).\\
Regarding the landscape attributes, they are learned using the SUN Attributes database~\cite{patterson2014sun}, from which we have chosen $33$  attributes relevant for our task (Figs. \ref{fig:classbeauty} and \ref{fig:templates}).

\begin{table}[]
    \centering
    \begin{tabular}{c|c|ccc}
    \hline
          & & \multicolumn{3}{c}{\# images}\\
         Data & Label & Training & Validaiton & Test \\\hline
         SoN& $\mathcal{Y}$ & $150'000$ & $20'000$ & $30'000$ \\
         SUN& $\mathcal{A}$ & $11'414$ & $1'497$ & $1'429$ \\
         Son + SUN & $\mathcal{A,Y}$ & - & - & $90$\\
         \hline
    \end{tabular}
    \caption{Number of samples per dataset}
    \label{tab:n}
\end{table}

In addition to checking the scenicness estimation performances of our model, we need to verify that the attributes are being correctly predicted on the images from the SoN database. To this end, an additional set of 90 SoN images was labeled by 4 different annotators with the same 33 attributes from SUN. Note that this set of SoN images with SUN attributes is used for validation purposes only and it is never using during training of either block of SIAM.





\subsection{Performance on the original datasets}

Table~\ref{tab:full} shows the impact of using a constrained-but-interpretable representation bottleneck on the  performance in both tasks. We report both the Root Mean Square Error (RMSE), as well as Kendall's $\tau$ rank correlation coefficient, as in~\cite{seresinhe2017using}, to assess the performances on the scenicness estimation task. Average precision is reported for the attribute detection task~\cite{patterson2014sun}. 

As a baseline for SoN, we use a finetuned ResNet-50, pretained on ImageNet, on the task of regressing scenicness values without making use of attributes. This results in a performance comparable to the one reported in~\cite{seresinhe2017using}, where the authors obtained values for Kendall's $\tau$ ranging from $0.62$ to $0.65$ on their test set using finetuned DL models.

\begin{table}[h!]
	\centering
	\begin{tabular}{r|c|c|c|}
		\cline{2-4}
		& SoN RMSE & SoN $\tau$ & SUN \\ \hline
		\multicolumn{1}{|l|}{Baseline } & 0.987 & 0.640 & -  \\ \hline
		\multicolumn{1}{|l|}{SIAM (ours)} & 1.01 & 0.607 & 0.418 \\ \hline
		\multicolumn{1}{|l|}{SIAM (no finetuning)} & 1.24 & 0.496 & 0.331 \\ \hline
	\end{tabular}
\caption{Numerical results. Scenicness prediction on the SoN test set (RMSE and Kendall's $\tau$) and SUN attribute prediction on the SUN dataset (average precision). The last row corresponds to SIAM with only $g$ trained on SoN, instead of the full model $f\circ g$.}
\label{tab:full}
\end{table}

We observe that training our model SIAM in two separate steps, $f$ only on SUN and then $g$ only on SoN with $f$ frozen, results in a substantial drop in accuracy, with a $25\%$ increase in RMSE with respect to the baseline. However, finetuning the whole model, $g\circ f$, jointly on SoN and SUN, not only reduces substantially this gap (by an order of magnitude, to $2.3\%$ RMSE increase), but also significantly improves the prediction on the subset of SUN attributes: the average precision on the SUN test set increases from $0.33$ to $0.41$.
This suggests that both tasks are correlated and confirms that optimizing jointly over the two losses does not penalize the attribute detection. This is of high importance for the final interpretability of the model, since a substantial drop in the attribute detection performance would risk making the interpretation of each attribute map meaningless. 

\subsection{Attribute detection on ScenicOrNot images}

The small set of 90 SoN images, selected to represent the whole range of scenicness values and annotated with SUN attributes, with 10 images randomly selected for each bracket between integer scenicness scores. This allows us to get an idea of the co-occurrence between the presence of these classes and the scenicness values, as shown in Fig.~\ref{fig:classbeauty}. We can see how a few classes, such as those related to water (\emph{ocean}, \emph{still water} and \emph{running water}), \emph{rugged} and \emph{hiking} are very correlated with high scenicness values, while most man-made classes tend to co-occur with below average scenicness. A few other classes, such as \emph{trees}, \emph{grass}, \emph{clouds} or \emph{farming} are much less polarized in terms of their average  associated scenicness.

\begin{table}[h!]
	\centering
	\begin{tabular}{r|c|c|c|}
		\cline{2-4}
		& SoN RMSE & SoN $\tau$ & SUN \\ \hline
		\multicolumn{1}{|l|}{Baseline } & 1.23 & 0.747 & -  \\ \hline
		\multicolumn{1}{|l|}{SIAM (ours)} & 1.22 & 0.700 & 0.501 \\ \hline
		\multicolumn{1}{|l|}{SIAM (no finetuning)} & 1.68 & 0.526 & 0.449 \\ \hline
	\end{tabular}
\caption{Numerical results on the 90 images SoN subset (SoN+SUN in Table~\ref{tab:n}). Scenicness prediction on the SoN test set (RMSE and Kendall's $\tau$) and SUN attribute prediction on the SUN dataset (average precision). The last row corresponds to SIAM with only $g$ trained on SoN, instead of the full model $f\circ g$. The average agreement between the four annotators on the attributes is 0.496.}
\label{tab:sunson}
\end{table}

Table~\ref{tab:sunson} shows the numerical results on these 90 images. The models used here, including the baseline, are the same described in the previous section and were trained on the original training sets of SoN and SUN (Table~\ref{tab:n}). The over-representation of images with very low and very high scores favours both higher RMSE and Kendall's $\tau$ values.
On scenicness prediction, SIAM matches the performance of the unconstrained model in terms of RMSE, although still lags behind in terms of Kendall's $\tau$.
We observe a substantial improvement on the SUN attributes detection task, matching the average agreement between annotators, which is of $0.496$. This suggests that the improvement on SUN attributes prediction observed in the SUN database (Table~\ref{tab:full}) generalizes well to SoN images, which do not contain SUN attribute labels at training time. This indicates that the semantic bottleneck can be indeed trusted when used to interpret the contribution of each attribute to the scenicness on the SoN images.

\subsection{Visual analysis}

\paragraph{Attribute templates.}
Figure~\ref{fig:templates} shows the per-attribute SIAM templates $\mathbf{T}^i$ learned  for predicting the scenicness. Dark green represents a large positive impact on the scenicness when the attribute is present in the corresponding location, while dark magenta means a large negative contribution. White represents a null contribution. The templates in Fig.~\ref{fig:templates} show that the attributes that contribute most consistently are, on the positive side, \emph{hiking}, \emph{rugged}, water (\emph{ocean} and \emph{still}) and \emph{ice} (which includes snow), and on the negative side \emph{metal}, \emph{glass}, \emph{wire}, \emph{transport} and \emph{dry}. The remaining attributes show some level of location dependency (\emph{e.g.} \emph{farming} has some positive impact when in the bottom half of the image but a negative one when it is located on the top third of the image) but have an overall weaker impact on scenicness.

\paragraph{Activation maps.}
We analyze the images from the SoN test set in which our model and the baseline disagree the most. For each image, we show the eight predicted attribute maps that contribute the most to the scenicness, both positively (with a green frame) and negatively (with a magenta frame). The thickness of the frame around each map represents the magnitude of the contribution to the scenicness score. For the baseline and the proposed model we show the total activation maps, which show how the contributions are distributed spatially.

\begin{figure*}[h]
    \centering
    \includegraphics[width=1.9\columnwidth]{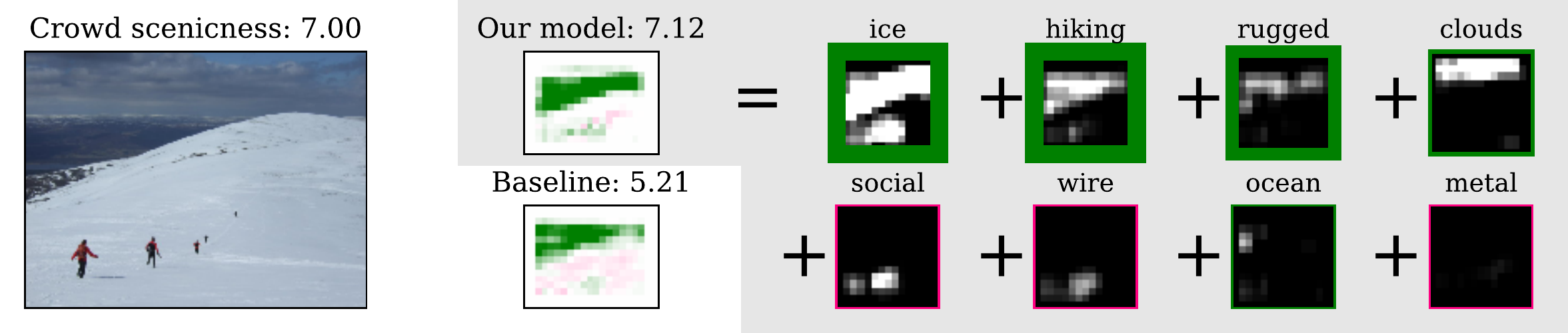}
    \includegraphics[width=1.9\columnwidth]{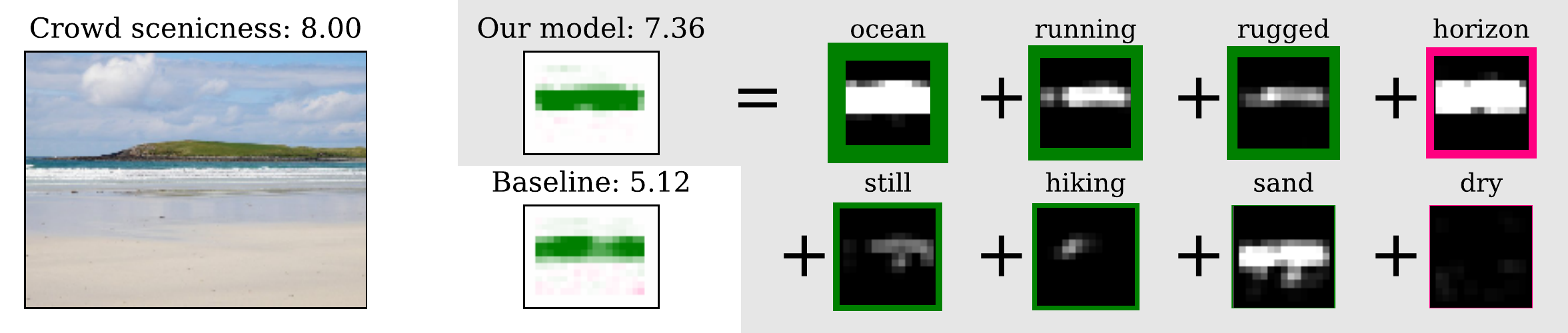}
    \includegraphics[width=1.9\columnwidth]{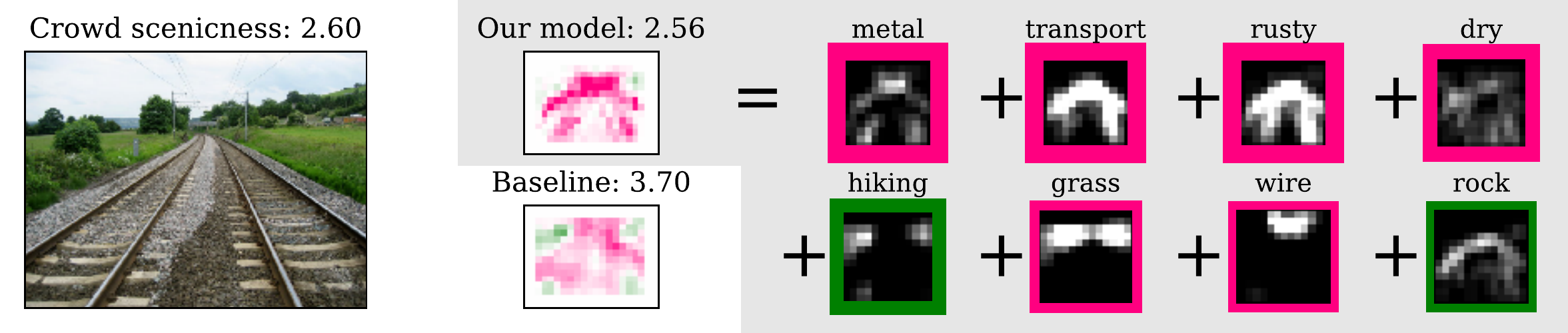}
    \caption{Examples in which SIAM predicts well both attributes and scenicness.}
    \label{fig:im1}
\end{figure*}

Figure~\ref{fig:im1} illustrates some examples where our proposed model (SIAM) performs well in attribute detection. The spatial distribution of the scenicness is similar in both SIAM and the baseline, but the elements that induce the latter to fail are not straightforward to discern, showcasing how explanations by localization might not always be satisfactory.

\begin{figure*}[h]
    \centering
    \includegraphics[width=1.9\columnwidth]{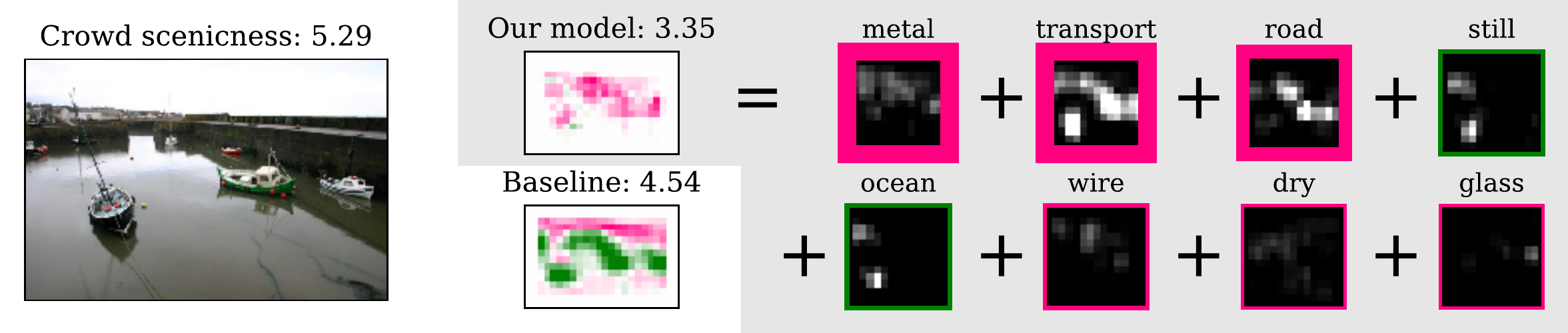}
    \includegraphics[width=1.9\columnwidth]{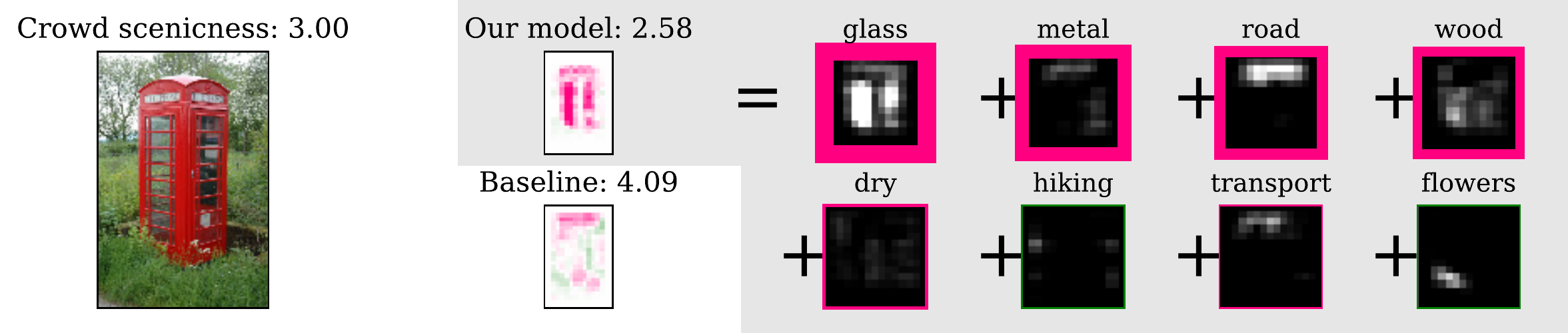}
    \includegraphics[width=1.9\columnwidth]{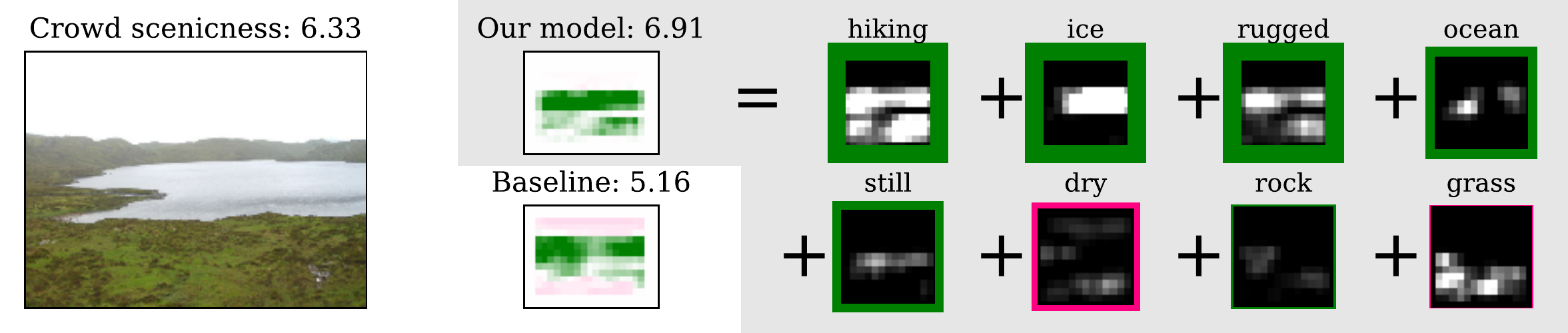}
    \caption{Examples in which SIAM mispredicts scenicness, but where this can be easily detected using the attribute prediction. }
    \label{fig:im2}
\end{figure*}

\begin{figure*}[h]
    \centering
    \includegraphics[width=1.9\columnwidth]{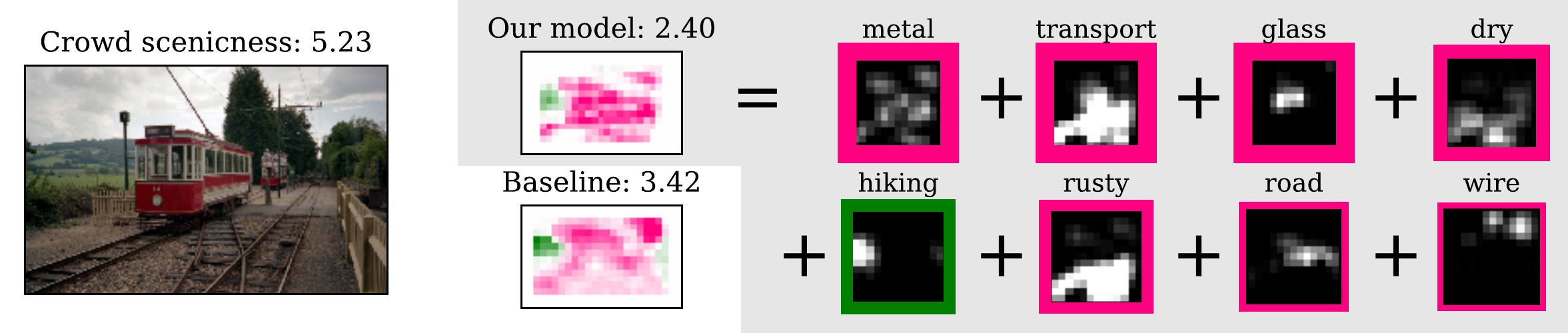}
    \includegraphics[width=1.9\columnwidth]{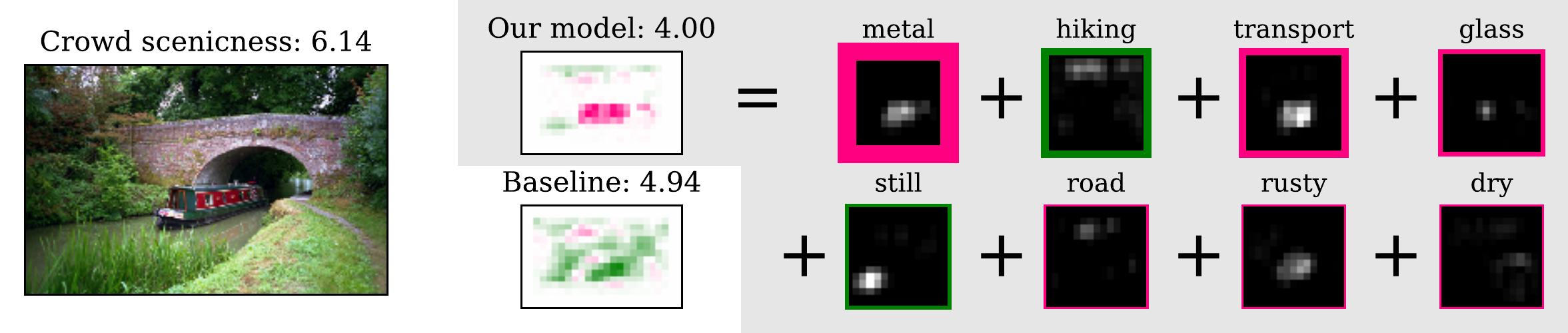}
    \includegraphics[width=1.9\columnwidth]{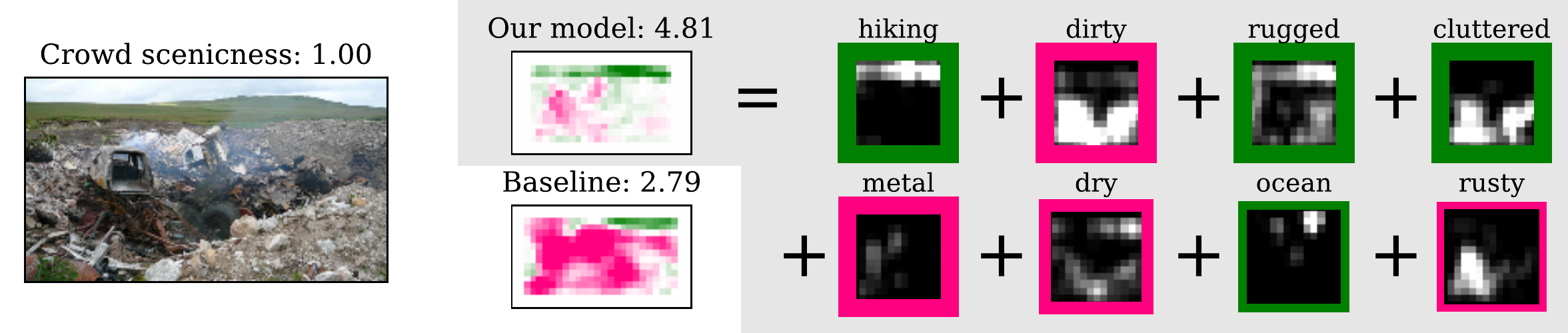}
    \caption{Failure cases in which SIAM mispredicts the scenicness without obvious mistakes on attributes prediction. }
    \label{fig:im3}
\end{figure*}

Figure~\ref{fig:im2} shows cases in which the error in scenicness can be easily attributed to misclassifications at the attribute level, allowing to correctly guess whether the model is over- or underestimating the score. In the top example, the reflection in the water is misclassified as road, impacting the prediction negatively. In the middle, the top of the phone booth is also misclassified as road, driving down the estimation of the score. In the bottom case, the reflection on the lake is predicted as ice, which is assigned a large positive contribution.

The examples in Fig.~\ref{fig:im3} represent cases where the baseline model captures  subtleties related to attributes (or object classes) that are not explicitly considered. In this case, SIAM remains blind to those contributions, since it is constrained to use only the pre-selected classes in the interpretable bottleneck.  In the top and middle examples, the attribute classes \emph{metal} and \emph{transport} are not able to capture subtleties such as the added aesthetic value of a vintage tram or a pleasant boat. In the last one, the positive scores of the \emph{rugged} and \emph{mountain} features overwhelm the dirtiness detected in a landfill.


These examples show that a fix and predefined set of attributes might not suffice, and a method for the discovery of potentially useful attributes could play a role in the selection of additional attribute classes. In addition to this, we often see that, although reasonable, the activation maps do not always match well with the semantics of the image. This is due to the weakly supervised nature of the attribute learning process, and some additional supervision, in the form of segmentation maps, could help solve this issue.

\section{Conclusion}
We propose the use of a semantic bottleneck made of Semantically Interpretable Activation Maps (SIAM) to provide an explanation of a CNN's output. These maps inform about what objective elements are relevant, where in the image they are, and how they contribute to the final prediction. We applied this method to the subjective task of landscape scenicness estimation, by forcing the model to use an information bottleneck that is jointly trained to predict a set of 33 landscape related attributes from the SUN Attributes database. 
Firstly, looking at the model layers that use the attribute maps as input, we can understand how the output will react to the presence of a given class at a given location on the images. Secondly, when an image is shown to the model, the activation maps and their contribution to the final score can help the user understand which elements are being used to construct the final score and get a hint about potential sources of errors.
Despite a small loss of performance (smaller than $2.5\%$ in terms of RMSE) in scenicness estimation, we observed a boost in the attribute detection and, more importantly, a much richer source of interpretation of the predicted value, without needing additional annotation. 


{\small
\bibliographystyle{ieee}
\bibliography{egbib}
}

\end{document}